\documentclass[letterpaper, 10 pt, conference]{ieeeconf}  % Comment this line out if you need a4paper

\IEEEoverridecommandlockouts                              % This command is only needed if 
\usepackage{graphicx}
\usepackage{mathptmx} % assumes new font selection scheme installed
\usepackage{times} % assumes new font selection scheme installed
\usepackage{amsmath} % assumes amsmath package installed
\usepackage{amssymb}  % assumes amsmath package installed
\usepackage{algorithm}
\usepackage{algpseudocode}
\usepackage{color}
\usepackage[dvipsnames]{xcolor}
\usepackage{romannum}

\title{\LARGE \bf
AOSoar: Autonomous Orographic Soaring of a Micro Air Vehicle
}

\author{Sunyou Hwang$^{1}$, and Bart D. W. Remes$^{1}$, Guido C. H. E. de Croon$^{1}$% <-this % stops a space
\thanks{$^{1}$ All authors are with the MAVLab, Department of Control and Operations, Faculty of Aerospace Engineering, Delft University of Technology, 2629HS Delft, the Netherlands
(email: {\tt\small S.Hwang-1@tudelft.nl, B.D.W.Remes@tudelft.nl, G.C.H.E.deCroon@tudelft.nl})%
}
\thanks{This paper has a supplementary video of flight tests and a replay of the flight log available at http://ieeexplore.ieee.org, provided by the authors.}
}

\begin{document}

\maketitle
\thispagestyle{empty}
\pagestyle{empty}

%%%%%%%%%%%%%%%%%%%%%%%%%%%%%%%%%%%%%%%%%%%%%%%%%%%%%%%%%%%%%%%%%%%%%%%%%%%%%%%%
\begin{abstract}

Utilizing wind hovering techniques of soaring birds can save energy expenditure and improve the flight endurance of micro air vehicles (MAVs). Here, we present a novel method for fully autonomous orographic soaring without a priori knowledge of the wind field. Specifically, we devise an Incremental Nonlinear Dynamic Inversion (INDI) controller with control allocation, adapting it for autonomous soaring. This allows for both soaring and the use of the throttle if necessary, without changing any gain or parameter during the flight. Furthermore, we propose a simulated-annealing-based optimization method to search for soaring positions. This enables for the first time an MAV to autonomously find a feasible soaring position while minimizing throttle usage and other control efforts. Autonomous orographic soaring was performed in the wind tunnel. The wind speed and incline of a ramp were changed during the soaring flight. The MAV was able to perform autonomous orographic soaring for flight times of up to 30 minutes. The mean throttle usage was only 0.25\% for the entire soaring flight, whereas normal powered flight requires 38\%. Also, it was shown that the MAV can find a new soaring spot when the wind field changes during the flight.

\end{abstract}

%%%%%%%%%%%%%%%%%%%%%%%%%%%%%%%%%%%%%%%%%%%%%%%%%%%%%%%%%%%%%%%%%%%%%%%%%%%%%%%%
\section{INTRODUCTION}

Flight endurance is one of the major factors holding back the real-world application of micro air vehicles (MAVs). % mavs repeating?
For low size, weight, and power (SWaP) MAVs, flight endurance mainly depends on the power density of the battery, which is limited \cite{elbanhawi2017enabling, traub2011range}, without fundamental progress on the horizon. 
%One way to improve flight endurance is to improve battery technology, but this development currently seems to progress slowly .
%There are ways to improve the stored energy such as combining batteries with hydrogen \cite{de2021nederdrone}, however, this is often not feasible for MAVs due to their size and weight limitations.
One way to improve flight endurance is to exploit energy from the environment.
Birds like albatrosses, vultures, ospreys, and kestrels are well known for their ability to actively use the wind to minimize their energy expenditure to fly longer distances or time \cite{bildstein1987hunting,strandberg2006wind, harel2016adult}. 
For example, vultures utilize energy from rising air columns created by uneven ground heating, called thermals \cite{pennycuick1971gliding}. Thermal soaring of unmanned aerial vehicles (UAVs) has been studied in various aspects, not only through manually developed guidance and control strategies \cite{allen2007guidance, edwards2008implementation} but also through reinforcement learning of detecting and exploiting thermals \cite{dunn2012unmanned, reddy2018glider}.

Another type of soaring is orographic soaring. Kestrels are often observed hovering at a position over a dune without flapping their wings, which is called wind-hovering \cite{videler1983intermittent}. This is a good example of the orographic soaring, 
%which is a specific type of static soaring, 
using the updraft generated by obstacles such as hills, mountains, and buildings when the wind hits the obstacle. 
%\textcolor{orange}{ In order to better understand the wind-hovering flight, Penn et al. developed an indoor environment replicating the kestrel's soaring condition in a wind tunnel \cite{penn2022method}. Wind-hovering flight tests of kestrels were conducted in the facility and precisely monitored using a motion capture system. Using the flight test data, the kestrel's hovering locations with flow velocity, angle and turbulence conditions were investigated. }
Wind-hovering can be useful for remaining in a single place for observation, but also for prolonging the flight range. For example, gulls appear to plan their path to exploit orographic soaring on the way to their destination to save energy \cite{shepard2016fine, williamson2020bird}. In this paper, we solely focus on orographic soaring.
%Orographic soaring is useful for surveillance applications (change the sentence). It can save a lot of time and energy by staying airborne for longer, since take off and landing are the most energy-inefficient parts of a flight. %(and urban environment)

% There are two types of soaring, dynamic soaring and static soaring.
% Orographic soaring is a specific type of static soaring, using uplift generated by obstacles such as hills, mountains, and buildings. (orographic uplift; when the wind hits an obstacle, uplift occurs)
% Unlike thermals that move depending on the weather and time, 
% Orographic updraft is useful because thermals move depending on the weather and ground conditions but orographic updraft is generated from obstacles so it stays at the same place. Especially in the urban environment, tall buildings generate orographic updraft that can be used for the soaring. (ref)
% often relatively small compared to the thermals

\begin{figure}[tb]
\centering
 \includegraphics[width=1.0\linewidth]{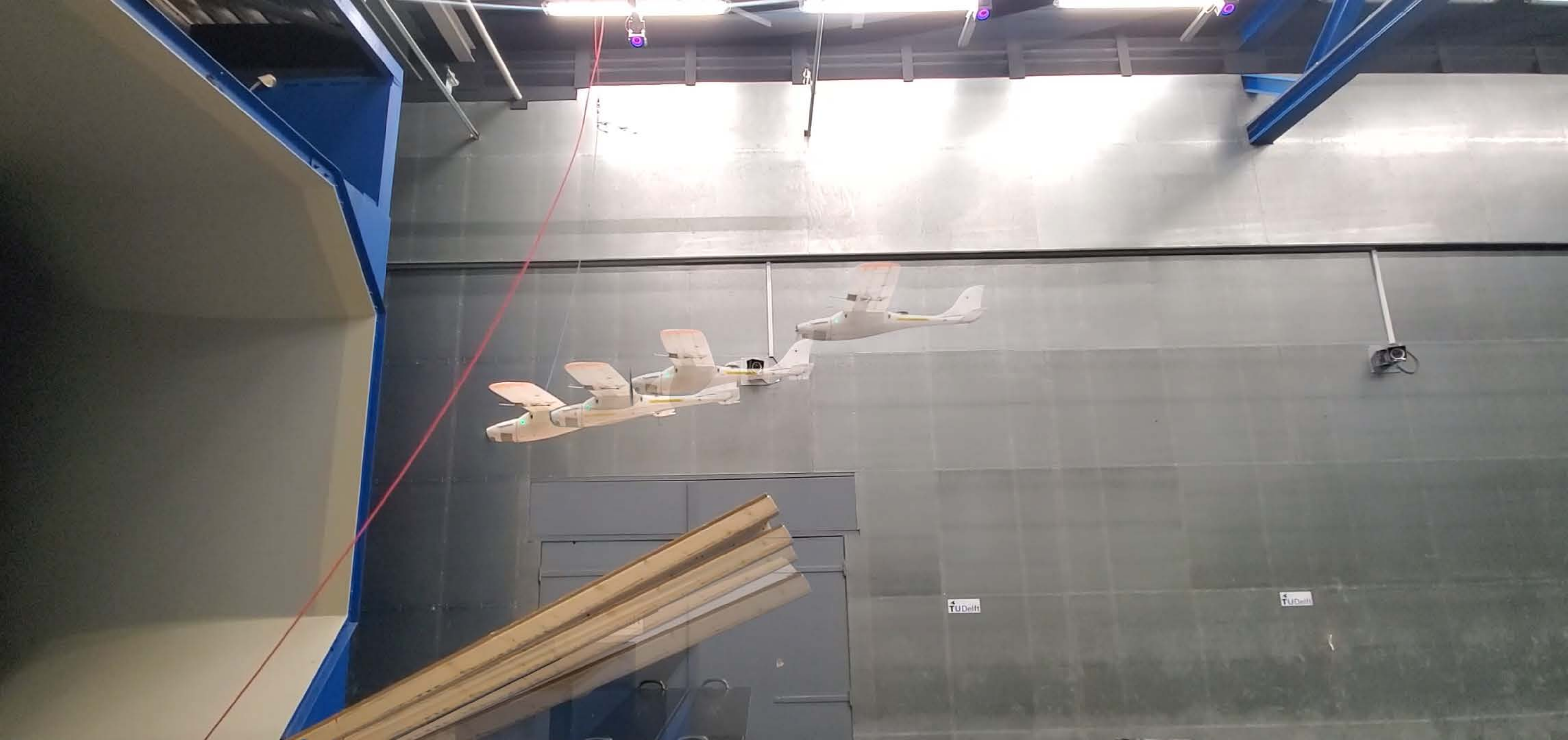}
 
 \caption{Autonomous orographic soaring of a micro air vehicle. Without a priori knowledge of the wind field, the MAV successfully performs autonomous soaring in the wind tunnel using little to zero throttle throughout the flight. The MAV can find a new position to soar when environmental conditions vary without any human intervention or manual parameter changes. In this picture, the increasing slope angle changes the wind field, and the MAV autonomously finds a new soaring position. It moves to the front and downward because of the combination of changes in the updraft and the MAV's sink rate.}
 
 \label{f:cover_soaring}
\end{figure}

%% previous research
%
%\textcolor{gray}{ analyzing orographic soaring condition (wind field, updraft, turbulence) in only simulation
% autonomous soaring flight test was not demonstrated}

%
%
Wind fields around obstacles and flight conditions for orographic soaring were analyzed using simulations and measurement data \cite{penn2022method,white2012feasibility, white2012soaring,guerra2020unmanned}.
These studies introduced unique opportunities and possibilities for exploiting orographic updrafts using MAVs. The feasibility and strategies of orographic soaring were also discussed. However, there was no actual flight demonstration performed in these studies.

Orographic soaring of MAVs in a real-world environment has been demonstrated in only a few studies \cite{fisher2015emulating, de2021never}.
However, the method in \cite{fisher2015emulating} was based on a priori knowledge of the entire wind field to generate a pre-defined trajectory to the soaring spot, while the method in \cite{de2021never} required manually positioning the MAV at a precise initial soaring position before switching on the autonomous soaring controller. 
%Fisher et al. demonstrated a soaring flight with a glider MAV with as a goal to gain altitude while maintaining the horizontal position \cite{fisher2015emulating}. In this study, a priori knowledge of the wind field was assumed to generate a trajectory for MAV to follow for soaring flight.
%Orographic soaring in front of a moving ship was demonstrated by de Jong et al. \cite{de2021never}. A priori knowledge of the wind field was not required in this study, however, one limitation still remained; a human pilot had to manually locate and initialize the MAV at a soaring position before switching on the autonomous soaring controller.
To achieve a fully autonomous soaring, considerable challenges remain. 
%Previous studies showed the feasibility of orographic soaring. Although orographic soaring in the real-world environment has been demonstrated before, it was assumed that a human pilot manually steered the MAV to a feasible soaring position, or a priori knowledge of the wind field was provided. 
In practice, accurately predicting or measuring the wind field is not feasible. Moreover, the MAV has to be able to explore and look for a feasible soaring position autonomously.

In this paper, we demonstrate for the first time the autonomous orographic soaring of an MAV without a priori knowledge of the wind field nor precise initial positioning of the MAV by a human pilot.
To achieve this, we present (\romannum{1}) a local search algorithm to find a soaring position and (\romannum{2}) an INDI controller with control allocation that enables the MAV to use the same controller setting during the entire flight.
An important advantage of the proposed control method is that the MAV does not need to switch controllers between soaring and navigation and can use the throttle whenever necessary.
We demonstrate the proposed method with a real-world flight in a wind tunnel. Moreover, we validate the versatility of the proposed methods by changing the wind speed and updraft during the autonomous soaring flight.

The paper is structured as follows:
In section \ref{methods}, 
an INDI-based soaring controller and searching method are presented. In section \ref{hardware}, we introduce the MAV and the wind tunnel test setups.
In section \ref{results}, the results from an autonomous soaring flight in the wind tunnel are presented. 
We discuss the flight test in section \ref{discussion}. Finally, we draw conclusions and suggest future research directions in section \ref{conclusion}.

\section{Methods} \label{methods}
%\textcolor{gray}{There are many challenges for autonomous orographic soaring. In this article, we focused on two main aspects.
%The first one is the controller. Orographic soaring is only feasible in a small region of the wind field. Furthermore, high wind speed is required for wind hovering.}

There are many challenges for autonomous orographic soaring. In this article, we focused on two main aspects.

The first one is the controller. 
There is a unique challenge for orographic soaring because the aim is to maintain the position without using the throttle. Also, it is only feasible in a small updraft region of the wind field with high wind speed.
Previous studies used a glider plane without a motor or changed controller or gains when the MAV enters the soaring mode. However, for sake of control fluidity, it would be desirable to use a single controller.
We adopted INDI with control allocation to enable using a single controller during the whole flight, regardless of navigation or soaring. 
We present our control method in section \ref{indi}.

The second aspect is to find feasible soaring positions. 
For fully autonomous flight, the MAV should be able to find where it can soar.
In previous studies, it was determined either from a priori knowledge of the wind field or by a human pilot.
We present a method for the MAV to autonomously find feasible soaring positions based on simulated annealing in section \ref{sa}.

\subsection{INDI controller with allocation} \label{indi}
Traditional PID controllers were used in most previous research, however, many of the authors have mentioned the need for a more advanced controller, especially for gust rejection. Therefore, we adopted an INDI controller for soaring flights. INDI is good at disturbance rejection and requires little model information \cite{sieberling2010robust,smeur2016adaptive,smeur2020incremental}.
INDI is an incremental form of nonlinear dynamic inversion. It controls angular acceleration $\dot{\omega}$ in an incremental way. The only required knowledge is the control effectiveness $G$, mapping an increment in the control input $u$ to a resulting rotational acceleration increment $\dot{\omega} - \dot{\omega_0}$ : 
%The relationship between an increment in input and angular acceleration increment caused by it, and control law can be represented as follows:

\begin{equation}
    \label{eq:indicontrollaw}
    \begin{aligned}
    \dot{\omega} = \dot{\omega_0} + G (u-u_0) \\
    u = u_0 + G^{-1}(\nu-\dot{\omega_0})
    \end{aligned}
\end{equation}

%\begin{equation}
%    \label{eq:effectiveness}
%    \dot{\omega} = \dot{\omega_0} + G(u-u_0)
%\end{equation}

% ctrl effectiveness

Where $\nu$ is the virtual control vector, and subscript 0 indicates a time in the past.
The control effectiveness depends on the inertia of the vehicle. However, directly measuring the inertia can be challenging. Alternatively, it can be estimated from flight test data with actuator inputs with angular acceleration. 
Using the flight test data, the control effectiveness matrix(G) can be estimated by dividing angular acceleration by a control input vector($u$).
Practically, we conducted several manual outdoor flight tests to log radio control input commands and angular accelerations by post-processing the inertial measurement units reacting to the radio input. The control effectiveness was calculated by dividing the change of angular acceleration by the change of input commands from radio control, for each pitch, roll, and yaw axis at various airspeeds. After that, the effectiveness values were fitted as a quadratic function and scheduled by airspeed measurement, because the effectiveness of control surfaces depends on the dynamic pressure $q = \frac{1}{2} \rho V^2$. 

\begin{figure*}[bth]
\centering
 \includegraphics[width=\linewidth]{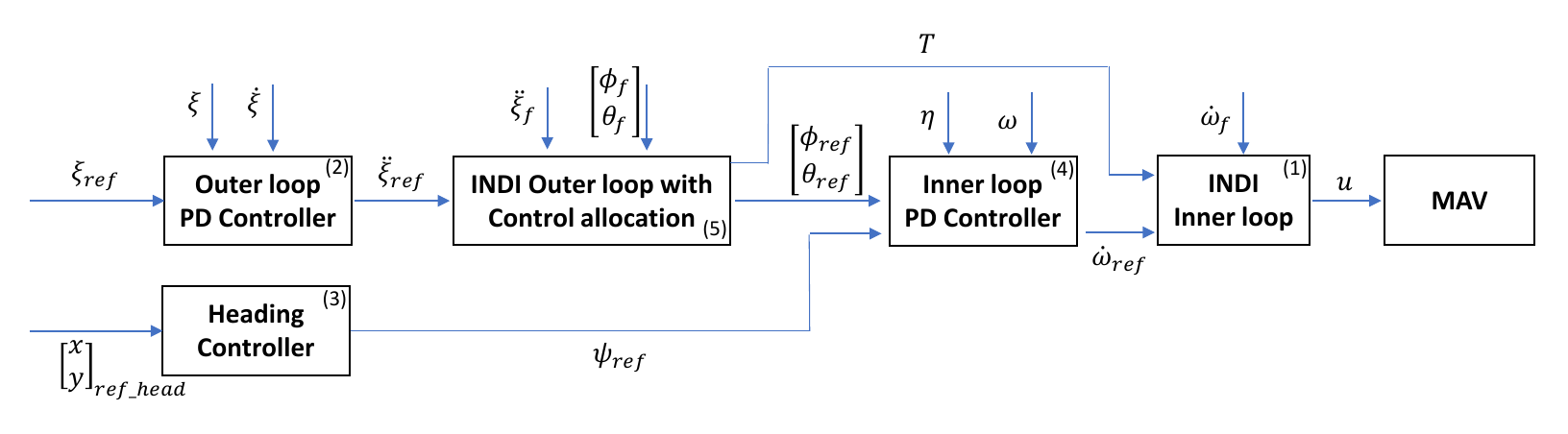}
 
 \caption{A schematic overview of the soaring controller. $\xi$ is the position, $\eta$ is the attitude, $\omega$ is the angular rate, $\psi$ is the yaw angle, $T$ is the thrust, and $u$ is actuator commands. Subscript $ref$ means reference, and subscript $f$ is for filtered signals.}
 \label{f:controller}
  % \vspace{-2mm}
\end{figure*}

%The throttle is useful to navigate to the soaring region or to deal with a strong gust. It is also necessary to use throttle when the MAV cannot fly without power due to the wind conditions. However, changing gains or switching the controller from one to another in flight to enable or disable the throttle is not desirable. We utilized control allocation to cope with this problem. 
%An INDI controller with control allocation allows the MAV to use the same controller and parameters throughout the flight, but seamlessly cut off or increase the throttle when desired. 
%In particular, this makes it unnecessary to change anything when switching between navigating, searching for a soaring spot, and soaring flight.
% \textcolor{magenta}{control allocation repetitive}

Figure \ref{f:controller} shows the overview of the soaring controller.
An INDI controller is used for both the inner loop and the outer loop.
For the outer loop, linear acceleration error is calculated from position error and fed into the INDI outer loop.

The MAV's position($\xi$), velocity($\dot{\xi}$), and a reference position are passed to a PD controller, and a linear acceleration reference($\ddot{\xi}_{ref}$) goes into the INDI outer loop \cite{smeur2020incremental}. Then, pitch, roll references, and thrust increment are calculated from the outer loop. 
$K_{\xi}$ and $K_{\dot{\xi}}$ are the gains for position error and velocity error, respectively.

\begin{equation}
\label{eq:outer}
\ddot{\xi}_{ref} = K_{\dot{\xi}}(K_{\xi}(\xi_{ref} - \xi) - \dot{\xi})
\end{equation}

To keep the heading towards the wind, we calculate a yaw($\psi$) reference. We set a virtual waypoint at 15 meters into the center of the cross-section of the wind tunnel. The yaw reference is calculated based on the error between the MAV's position and the virtual waypoint. Let $x_{ref}, y_{ref}$ be the x and y position of the virtual waypoint, and $x, y$ the longitudinal and lateral position of the MAV. Then the yaw reference is determined by the following equation:
\begin{equation}
\label{eq:yaw}
\psi_{ref} = \mathrm{atan}(\frac{y_{ref}-y}{x_{ref}-x})
\end{equation}

The attitude reference($\eta_{ref}$ = $[$$\phi_{ref}$ $\theta_{ref}$ $\psi_{ref}$$]^{T}$) is passed to an inner loop PD controller. Then, the angular acceleration reference($\dot{\omega}_{ref}$) calculated from the PD controller and the thrust increment are passed into the inner loop INDI controller. 
\begin{equation}
\label{eq:inner_pd}
\dot{\omega}_{ref} = K_{\omega}(K_{\eta}(\eta_{ref}-\eta)-\omega)
\end{equation}

The throttle is useful to navigate to the soaring region or to deal with a strong gust. It is also necessary to use throttle when the MAV cannot fly without power due to the wind conditions. However, changing gains or switching the controller from one to another in flight to enable or disable the throttle is not desirable. We utilized control allocation to cope with this problem. 
An INDI controller with control allocation allows the MAV to use the same controller and parameters throughout the flight, but seamlessly cut off or increase the throttle when desired. 
In particular, this makes it unnecessary to change anything when switching between navigating, searching for a soaring spot, and soaring flight.

Control allocation of INDI was originally developed to prevent actuator saturation for over-actuated vehicles, because of aggressive yaw control behaviour. The weighted least square (WLS) algorithm was integrated with the INDI controller for inner loop control allocation in a paper by Smeur et al. \cite{smeur2017prioritized}. 
Here, we use control allocation for achieving as still soaring as possible. This implies a preference for reducing throttle and minimizing accelerations. This leads to the following control allocation cost function: 
%The cost function for the input $C_{alloc}(u)$ is defined in equation \ref{eq:WLS}. We want to find a control input that minimizes the cost function $C_{alloc}(u)$.
\begin{equation}
\begin{aligned}
    \label{eq:WLS}
    C_{alloc}(u_{wls}) = \lVert W_u(u_{wls}-u_{p,wls}) \rVert ^2 + \\
    \gamma \lVert W_v(G_{o}u_{wls}-v_{wls}) \rVert ^2
\end{aligned}
\end{equation}
Where $u_{wls}$ is a control increment vector, %which is increment in [$\phi$ $\theta$ $T$].
$W_u$ is a weighting matrix for the control inputs, $W_v$ is a weighting matrix for the control objective, $G_{o}$ is the control effectiveness matrix for the outer loop, $v_{wls}$ is the virtual control increment command for the outer loop, $u_p$ is the preferred control increment vector, and $\gamma$ is a scale factor.

%\textcolor{magenta}{Details of the method are described in a paper by Smeur et al. \cite{smeur2017prioritized}.}
% \cite{harkegard2002efficient}

In this study, control allocation is used to prioritize the controls to minimize thrust for the outer loop controller. We set a higher priority to control pitch than the thrust by adjusting weight for the WLS optimization. 
%The MAV uses the throttle only if it cannot cope without increasing the throttle. \textcolor{orange}{Otherwise, the controller decreases the throttle usage by default.} 
Therefore, for $W_v$, we choose the weights to be 1, 100, 1 for roll($rad$), pitch($rad$), and thrust in a range of [0, 9600] respectively. For $W_u$, we choose 1 for all axes, $\gamma$ is $10^6$, and $u_p$ is a zero vector.

% Weight matrix, for each control objective and inputs (actuator)
% mainly throttle and pitch; G full rank?
% (roll, pitch, yaw, throttle); (aileron, elevator, rudder, thrust)
% outer loop allocation; position setpoint -> velocity -> linear acceleration; -> roll, pitch, thrust setpoint
% inner loop allocation; with effectiveness schedule (AS); roll, pitch, yaw, thrust setpoint -> actuator commands

\subsection{AOSearch: Autonomous Orographic Search for a soaring location} \label{sa}
To exploit updrafts, MAVs have to find feasible soaring locations autonomously. %has to know where is a feasible location for soaring. 
If the environment is static and a prior knowledge of the wind field is provided, it can be calculated from the MAV model and wind speed by finding an equilibrium. However, in the real world, it is difficult to measure the entire wind field. Furthermore, the wind speed may change during the flight. Thus, we developed an algorithm to find the soaring location which is applicable in a non-static environment, without any prior knowledge of the wind field.

% SA in general
Simulated annealing is a local meta-heuristic search technique \cite{kirkpatrick1983optimization, ingber2000adaptive}.
It attempts to minimize a cost function by taking steps to neighboring positions in the search space. It either accepts or rejects the solution based on an acceptance probability which usually decreases to zero as the "temperature" decreases. Initially, steps that increase the cost are allowed, but eventually, simulated annealing becomes a greedy algorithm.

% -temperature
% -acceptance probability
% -annealing schedule

% Energy function to be minimized is defined, % temperature decreases over time.
% E = fn(throttle, velocity, pitch rate)
% step decreased with E // decreasing over time is not appropriate because the environment can be changed during the mission (wind speed, slope angle)

% not necessary to find an optimal position but a feasible soaring position. --> threshold

Based on simulated annealing, we implemented \emph{AOSearch}: Autonomous Orographic Search algorithm.
In our case, there could be environmental changes at any time during the flight, regardless of the progress of the search. For example, wind speed can be changed over time. 
Therefore, the temperature does not decrease at each step. The temperature is set to zero, so the search algorithm only accepts better solutions. %\textcolor{orange}{Because of the shape of the search space, avoiding local minima is not crucial and finding a global optimum is not necessary.} 
When the MAV finds a position at which it can soar, the search is finished.
Hence, we implemented a threshold cost. If the value of the cost function is lower than the threshold at a certain position, it is considered converged and does not move to a new neighbour. When the value of the cost function exceeds the threshold because of an environmental change, the MAV restarts the search until it finds a new position that satisfies the threshold condition.
%but stays at the same position until the value of the cost function exceeds the threshold. 
%%% yes, in general local/global is not necessary unless we want to regenerate the battery 'efficiently', +wind field changes(uncertainty)

We want to minimize energy expenditure as well as stay in the same position as much as possible just like kestrels hovering without moving its head for observation. In our case, thrust is the primary source of energy consumption to minimize. Pitch rate is also contributing because controlling the elevator spends energy. Furthermore, we aim for wind-hovering, which means that the MAV maintains its position. So, we want to minimize both horizontal and vertical speed.
The cost function($C_{search}$) captures all requirements for still, orographic soaring, with the MAV able to keep its downward view as static as possible with minimum throttle usage, also meaning minimal position and pitch changes. 
Hence, it is a function of thrust($T$$[\%]$), horizontal and vertical ground speed($\dot{x}$, $\dot{z}$$[m/s]$), and pitch rate($\dot{\theta}$$[rad/s]$) with a gain for each parameters. 
\begin{equation}
\label{eq:costfn}
C_{search} = k_1T + k_2|\dot{x}| + k_3|\dot{z}| + k_4|\dot{\theta}|
\end{equation}
The gains were $k_1$=9.6, $k_2$=1.6, $k_3$=1.0, and $k_4$=10. The threshold value was set to 43.
%k1 0.1*T(0<=T<=9600) k2 1.6 k3 1.0 k4 10(rad?deg?), averaged for 5sec, thres
The gains and threshold value were determined empirically, by observing the values when the MAV was soaring in a stable manner.

% The step size is 0.3 $m$ when $E \geq 3 \times threshold$, 0.2 $m$ when $3 \times threshold > E \geq 2 \times threshold$, 0.1 $m$ when $2 \times threshold > E \geq 1.5 \times threshold$, and 0.05 when $1.5 \times threshold > E$.

%%%%%%%%%%%%%%%%%%%%
\emph{AOSearch}: the autonomous orographic search method is described in algorithm \ref{alg:sa}. 
First, calculate the cost function. Based on the value of the cost function, a step size($S$) is selected. If the value is lower than the threshold, the MAV stays at the current position($Pos(s_{new})$).
If the value of the cost function has increased compared to the previous position($Pos(s)$), go back to the previous position. If the value has decreased, keep the same direction. If it has just come back from a previous position, pick a new random neighbour.
To get a random neighbour $s_{new}$, pick a random direction($Dir(s_{new})$) among four direction vectors of [x, z]: forward[1, 0], backward[-1, 0], up[0, 1], down[0, -1]. Then, a new soaring position is:
$Pos(s_{new})$ $=$ $Pos(s)$ $+$ $Dir(s_{new})$ $\times$ $S$. 
Repeat the process until the cost function value becomes less than the threshold.
%%%%%%%%%%%%%%%%%%

The step size (S) is set to four steps depending on the value of the cost function. As the value gets lower, the step size decreases from 0.3 $m$ to 0.05 $m$. The logic behind this is that the MAV takes a bigger step to explore the wind field when the energy consumption at the current position is high. When the value of the cost function is low at the current position, the MAV tries to fine-tune its position. 

$$
\small {
S = \begin{cases}
			0.3, & \text{for $C \geq 3 \times threshold$}\\
            0.2, & \text{for $3 \times threshold > C \geq 2 \times threshold$}\\
            0.1, & \text{for $2 \times threshold > C \geq 1.5 \times threshold$} \\
            0.05, & \text{otherwise}
		 \end{cases}
		 }
$$

%%% Not really sure how to make this clear...
\begin{algorithm}[bht]
\caption{AOSearch}\label{alg:sa}
\begin{algorithmic}[1]
% \State Calculate cost function
\footnotesize
\State $C(s_{new})$ $\gets$ $k_1T + k_2|\dot{x}| + k_3|\dot{z}| + k_4|\dot{\theta}|$    \Comment{Calculate cost function}
\State S $\gets$ Calculate a step size
    
\If{$C(s_{new}) < threshold$}
\State Stay at the current position

\Else
    \If{$C(s_{new}) < C(s)$}
        \If {returned $is$ True}
            \State $Dir(s_{new}) \gets random$  \Comment{pick a random direction}
            % \State $s_{new} \gets neighbour(s)$
            \State $returned \gets False$
        \Else
            \State $Dir(s_{new}) \gets Dir(s)$  \Comment{keep the same direction}
            % \State $s_{new} \gets Dir(s) \times S$
        \EndIf
    
    \Else
        % \State $s_{new} \gets s$    
        \State $Dir(s_{new}) \gets - Dir(s)$    \Comment{go back to the previous position}
        \State $returned \gets True$
    \EndIf
    
    \State $Pos(s_{new})$ $=$ $Pos(s)$ $+$ $Dir(s_{new})$ $\times$ $S$
    \State Move to a new soaring position $Pos(s_{new})$
    
\EndIf

\end{algorithmic}
 % \vspace{-1mm}
\end{algorithm}
%%%
%%%%%%%%%%% redo

\section{Hardware and test setup} \label{hardware}
% In this section, an overview of the hardware and wind tunnel test setup in the wind tunnel is described.

\subsection{Eclipson model C 3d-printed model plane}
An Eclipson model C \cite{eclipson} airplane was used for the flight tests. It is a 3d-printed plane, which makes it easy to replace parts in case of a crash. It was printed with lightweight polylactic acid (LW-PLA) to reduce weight and increase aerodynamic performance. 
Unlike other studies that used a glider or a flying wing with flaps, we chose a 5-channel model plane to have more control. It has four servos for control surfaces (elevator, left aileron, right aileron, and rudder) and an electric motor at the front. Especially the rudder is useful to keep its heading toward the wind while maintaining its lateral position to stay in a small updraft region. The throttle is also necessary to navigate and fly safely in case of a strong gust or sudden environmental changes.
%%% Added a line to explain 5-channel thing

\begin{figure}[bt]
\centering
\vspace{0.1in}
 \includegraphics[width=1.0\linewidth]{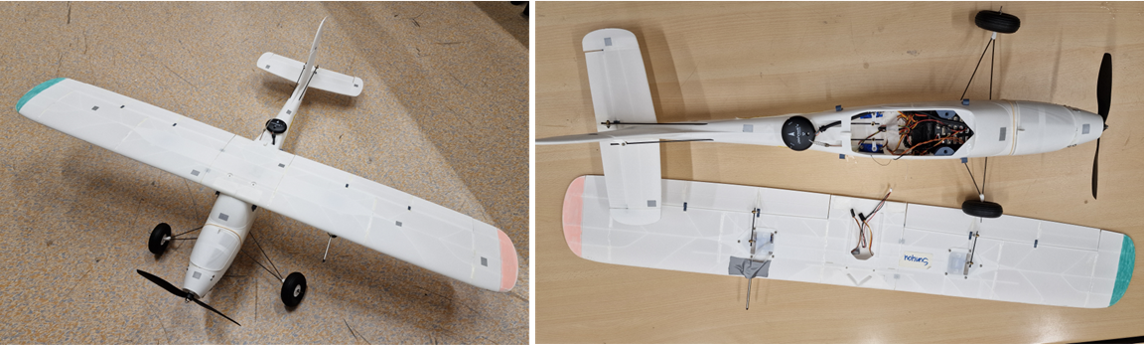}
 \caption{Eclipson model C 3D-printed airplane. It has 1.1$m$ wingspan and a weight of 716$g$ including a 1.5$A$ Li-Po battery. Pixhawk4 is equipped with Paparazzi open-source autopilot software. An airspeed sensor is mounted under the wing. GPS sensor and opti-track markers are used for localization.}
 \label{f:eclipson}
\end{figure}

The MAV is shown in Figure \ref{f:eclipson}. It has a wingspan of 1100$mm$, 18$dm^2$ wing surface area, and an aspect ratio of 6.9. A Pixhawk 4 with Paparazzi autopilot open-source software \cite{hattenberger2014using} was used. Pixhawk 4 board has a processor and an inertial measurement unit. A GPS module and opti-track markers were used for outdoor and indoor localization, respectively. An airspeed sensor was mounted under the wing and calibrated in the wind tunnel.
The weight of the aircraft with electronics was 595 grams excluding battery, and a total of 716 grams including a 1.5$A$ Lithium-Polymer battery.

\subsection{Wind tunnel test setup}
\begin{figure}[bt]
\centering
 \includegraphics[width=1.0\linewidth]{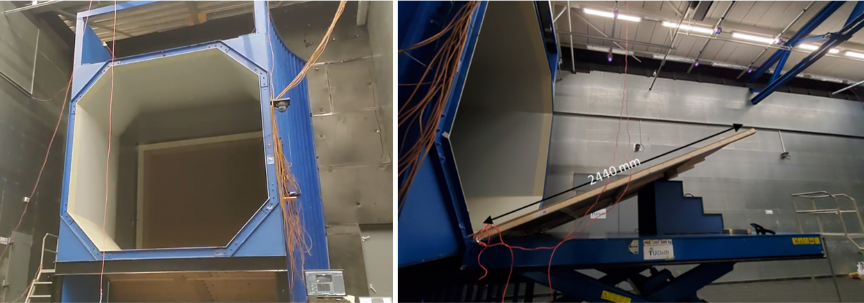}
 \caption{TU Delft OJF and the slope setting for generating orographic updraft. The cross-section of the wind tunnel is $2.85m\times2.85m$. A wooden plate ($2.44m\times2.44m$) was used for the slope. The slope was placed in front of the outlet of the wind tunnel.}
 \label{f:hardware_settings}
  \vspace{-2mm}
\end{figure}

\begin{figure}[bt]
\centering
 \includegraphics[width=1.0\linewidth]{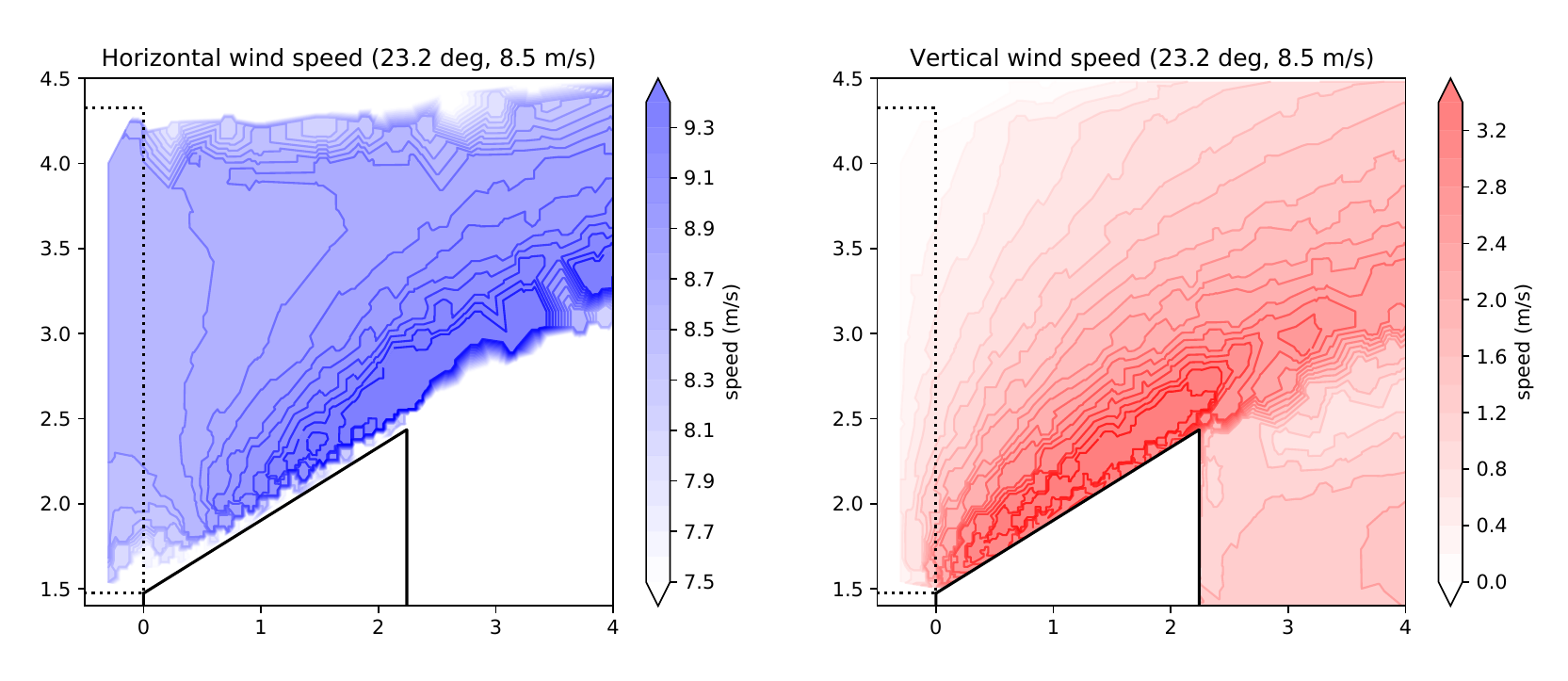}
 \caption{Contours of horizontal and vertical wind speed over a slope in the wind tunnel. The slope angle was set to 23.2 degrees and the wind speed from the OJF was set to 8.5 m/s.}
 \label{f:cfd_wind_speeds}
  % \vspace{-2mm}
\end{figure}

The TU Delft open jet facility (OJF) is a wind tunnel with a $2.85m \times 2.85m$ cross-section. We installed a ramp in front of the wind outlet to generate an orographic updraft. The OJF and the slope are shown in Fig. \ref{f:hardware_settings}. The slope angle and wind speed were adjustable during the flight. For safety reasons, a rope system was attached to the ceiling. An Opti-track system installed in the wind tunnel was calibrated before the flight test.
In order to get insight into the wind field with this setup, a computational fluid dynamics (CFD) simulation was performed using ANSYS fluent \cite{ansysfluent}. The contours of horizontal and vertical wind speeds are shown in Fig.\ref{f:cfd_wind_speeds}. The strongest updraft occurs close to the end of the slope, and the wind speed decreases near the slope because of the boundary friction.

\subsection{Glide polar}
\begin{figure}[bt]
\centering
 \includegraphics[width=0.9\linewidth]{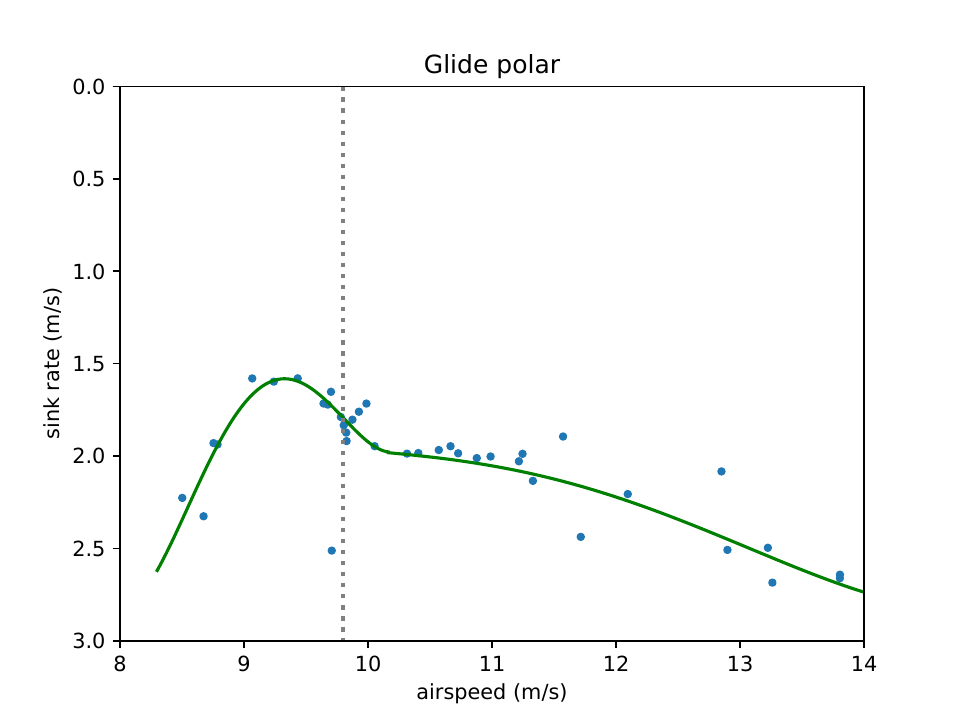}
 \caption{Glide polar of the Eclipson model C airplane. The data was divided into two sections and fitted with a fourth-order polynomial. Additional drag was generated at around 9.8 m/s of the airspeed because of the propeller windmilling.}
 \label{f:glide_polar}
  \vspace{-2mm}
\end{figure}

To measure the sink rate, a series of outdoor flight tests was performed. The MAV had the same hardware and weight setting with the indoor tests for consistency. We flew the MAV manually on a calm day, and retrieved sink rate and airspeed data from flight logs during parts of the flight where the motor was turned off. The data points and the glide polar are shown in Fig. \ref{f:glide_polar}. The data was divided into two sections and fitted with a fourth-order polynomial function. The sink rate sharply increases around 9.8 m/s of airspeed. This is because the propeller started windmilling at that airspeed, generating additional drag. 
% We did not use folding propeller because of practical difficulty with weight and shape of mounts. Instead, we used a normal 8.5x4 inch propeller.

\section{Test results} \label{results}

\begin{figure*}[tbh]
\centering
 \includegraphics[width=0.95\linewidth]{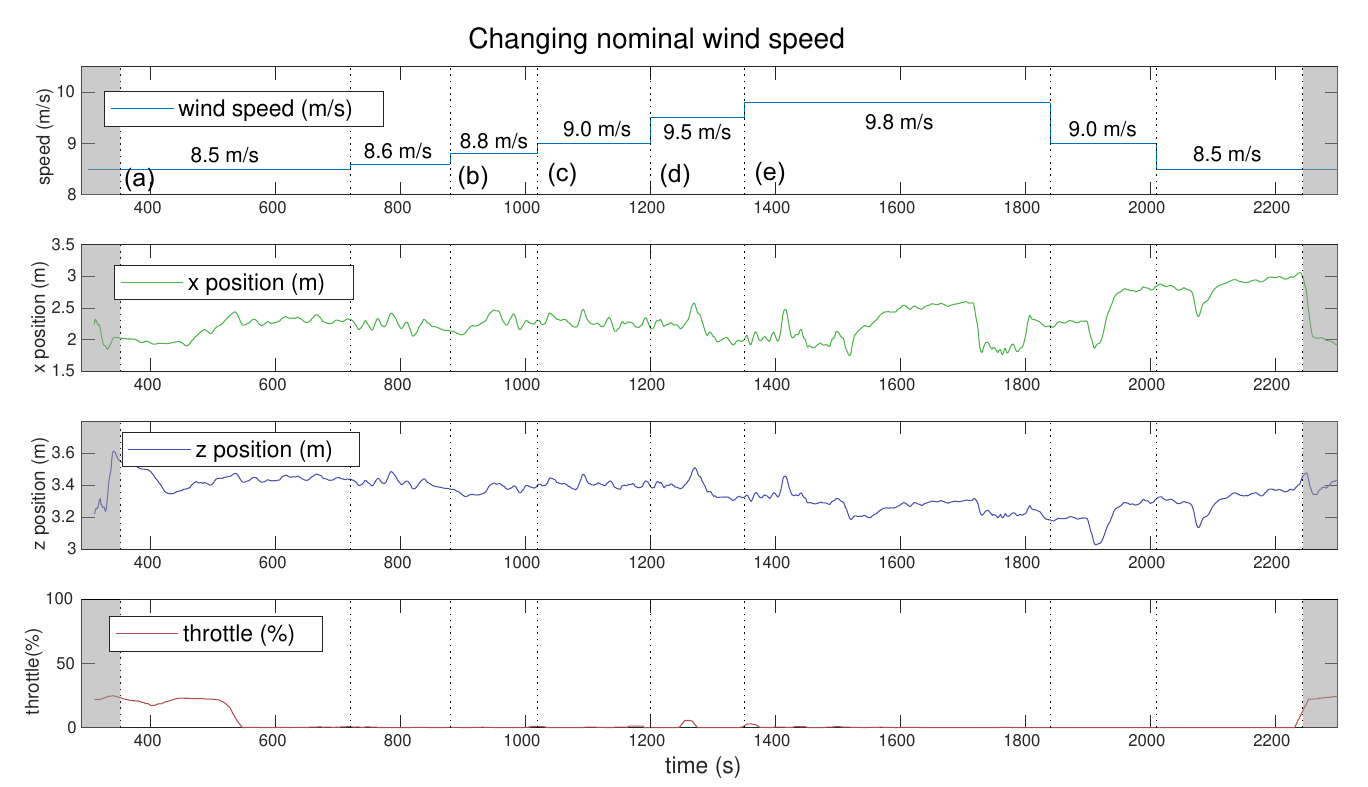}
 \caption{Horizontal and vertical position, and throttle usage of the MAV during the flight according to the change of the wind speed from the wind tunnel.}
 \label{f:plot_test_ws}
 \vspace{-3mm}
\end{figure*}

\begin{figure}[bth]
\centering
 \includegraphics[width=1.0\linewidth]{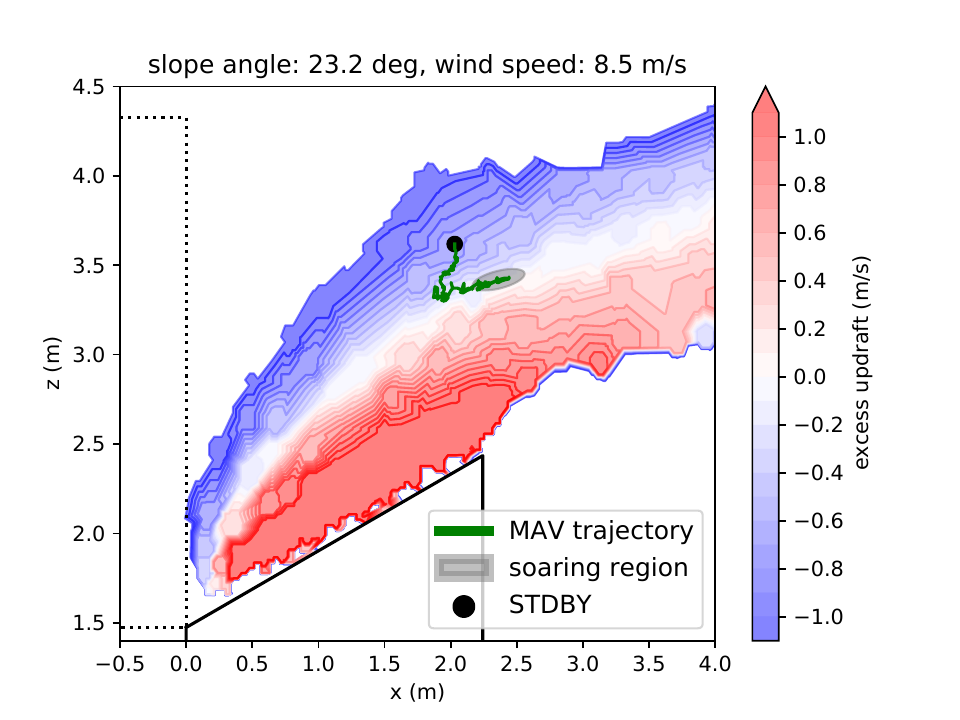}
 \caption{Flight trajectory during the autonomous search, from 352 to 537 seconds in case 1. The MAV started searching at the standby position (STDBY), marked as a black dot. It found a good soaring location and achieved zero throttle in 185 seconds.}
 \label{f:plot_searching_trajectory}
   \vspace{-3mm}
\end{figure}

\begin{figure*}[bth]
\centering
 \includegraphics[width=0.95\linewidth]{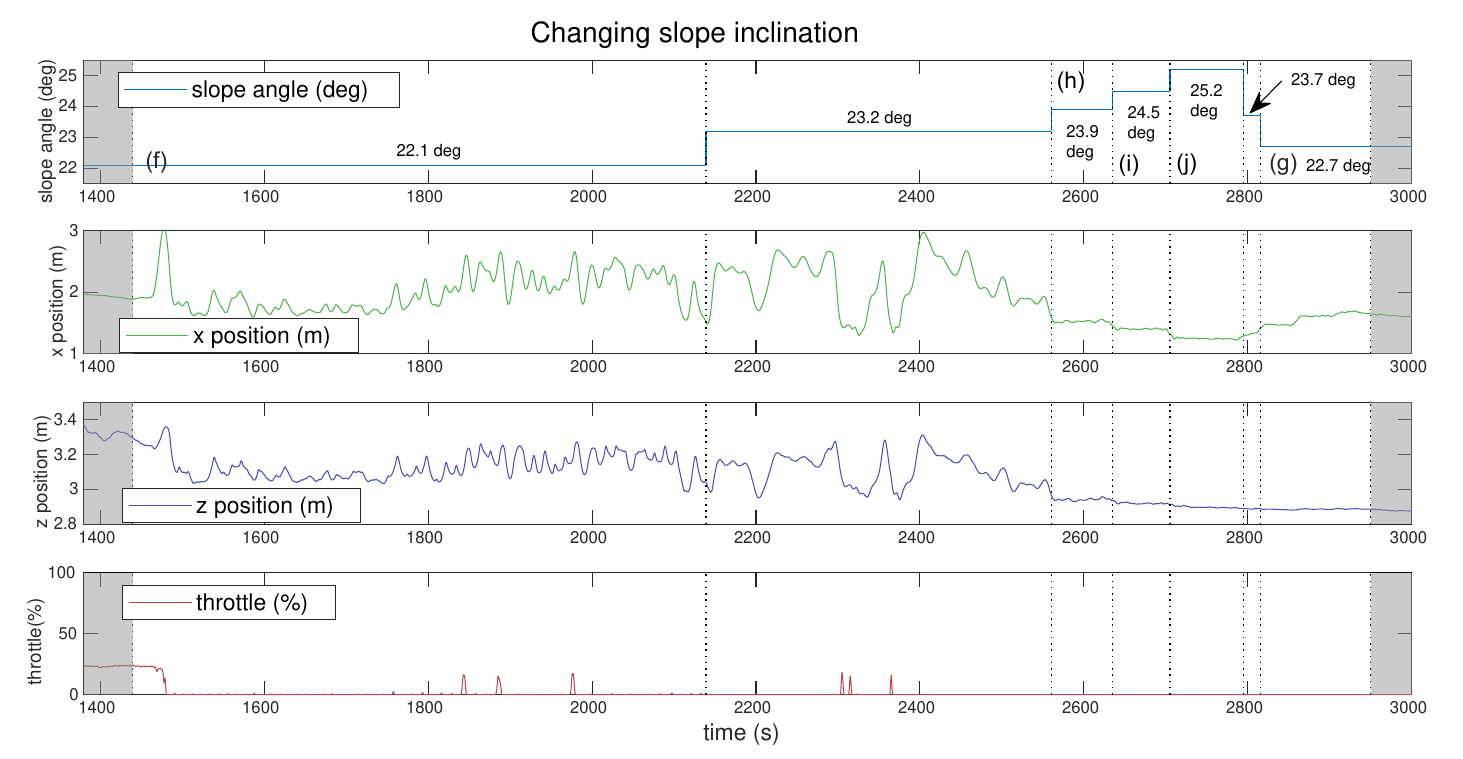}
 \caption{Horizontal and vertical position, and throttle usage of the MAV during the flight according to the change of the slope angle.}
 \label{f:plot_test_slope}
\end{figure*}

% cut 22.1~23.2?

\subsection{Autonomous Soaring}

% \subsection{Test procedure}
Indoor flight tests were performed to demonstrate autonomous soaring using the proposed methods in this paper. We first started the flight in manual mode for safety reasons until the nominal wind speed reached 8.0 $m/s$. Once the wind speed was stabilized, we switched on the autonomous flight mode. The MAV was fully autonomous from that point.
It first hovered at a standby position and stabilized itself. Then it started the local search and autonomously moved its target soaring position until it converged according to algorithm \ref{alg:sa}.
% \textcolor{blue}{The reference soaring position moved autonomously // and the MAV followed it.}
The standby position was set where requires approximately 20\% of the throttle for hover.
During the flight, we changed the environmental conditions, either the wind speed or the slope angle. %We waited sufficient time between each step.
Everything was running onboard, except position measurements from the opti-track. Data from onboard sensors was written on an SD card using a high speed flight logger and retrieved after the flight.

There were two experimental cases: one in which we changed the wind speed with a fixed slope angle, and another in which we changed the slope angle with a fixed wind speed. In both cases, the MAV was able to successfully soar using very little energy without any prior knowledge of the wind field. Furthermore, it was shown that autonomous soaring in a changing environment is possible by using the proposed methods.

\subsubsection{Changing nominal wind speed}
The first case is to change the wind speed during soaring. 
We started autonomous soaring at a wind speed of 8.5 m/s, and then slowly increased it to 9.8 m/s, and decreased it again.
When the wind speed changes, the MAV either stays at the same position if possible, or tries to find another position. If the current position becomes not feasible for soaring due to the change of the wind field, it restarts searching for a new feasible soaring position.

Figure \ref{f:plot_test_ws} shows the wind speed, horizontal and vertical position, and throttle usage during the flight. Figure \ref{f:plot_searching_trajectory} shows the trajectory during the autonomous search from an initial position at 8.5 $m/s$ wind speed. The throttle started of at 20\%, but after the autonomous search for a soaring location turned on at 352 seconds, the MAV found a good soaring location in 185 seconds, achieving zero throttle at 537 seconds. From that time on, the throttle usage was very low. %There were a few times the MAV used the throttle during soaring (1257~1264 15\%, 1175~1177 12\%, 1358~1364 6\%).
At 2244 seconds, we retook manual control to make the MAV land. Manual control is shown by the shaded regions.
The wind speed was changed over time, which led to changes in position. There is no clear proportional relationship between wind speed and the positions. We will analyze the chosen soaring positions further in the next section.
The flight time was a total of 30 minutes excluding a short manual flight after launch and landing. 
%The throttle usage decreased to 0\% after the soaring mode was switched on. 
During the soaring, the mean throttle usage was 0.25 \%. Note that it is a significant decrease in throttle usage compared to non-soaring flights. 38\% of the throttle was required for the MAV to hover in the wind tunnel without a ramp at 8.5 m/s wind speed, and 30 to 50\% of the throttle was normally used during outdoor flights.

\subsubsection{Varying slope angle}
In the second case we changed the slope angle during soaring. The procedure remained the same as the first case, but instead of changing the wind speed, we changed the slope inclination while the nominal wind speed was fixed at 8.5 m/s. The slope angle, the MAV's horizontal and vertical position, and the throttle usage are shown in figure \ref{f:plot_test_slope}.
The slope angle was set from 22.1 degrees to 25.2 degrees. The step size of the slope angle was not consistent because of the practical difficulty of moving the ramp precisely during the flight. The throttle usage was 25\% at the start. 
We started the autonomous search at 1440 seconds. The MAV achieved zero throttle flight at 1572 seconds, 132 seconds after starting the search. At 2950 seconds, we retook manual control for landing. 
With a slope angle of 22.1 degrees, the inclination was small, so less updraft was generated than other conditions. Because of that, the combination of the horizontal and vertical wind was not favorable to hover stationary.
Nevertheless, it was still able to soar using zero throttle at 22.1 degrees slope angle, allowing some movement.
The soaring position moved frontward when the slope inclination increased, and it moved backward again when the slope angle decreased. This is because the feasible soaring region pushed forward as the slope angle increased. We will analyze the change of the wind field and chosen positions in the next section.
The soaring flight time was a total of 25 minutes, excluding short manual flights after launch and before landing. During the soaring, the mean throttle usage was 0.25 \%. 
Note that in both test cases, the flight was stopped before using up all the battery because of the size of the flight log file and limited onboard memory.

\section{Analysis and Discussion} \label{discussion}
\begin{figure}[bthp]
\centering
 \includegraphics[width=1.0\linewidth]{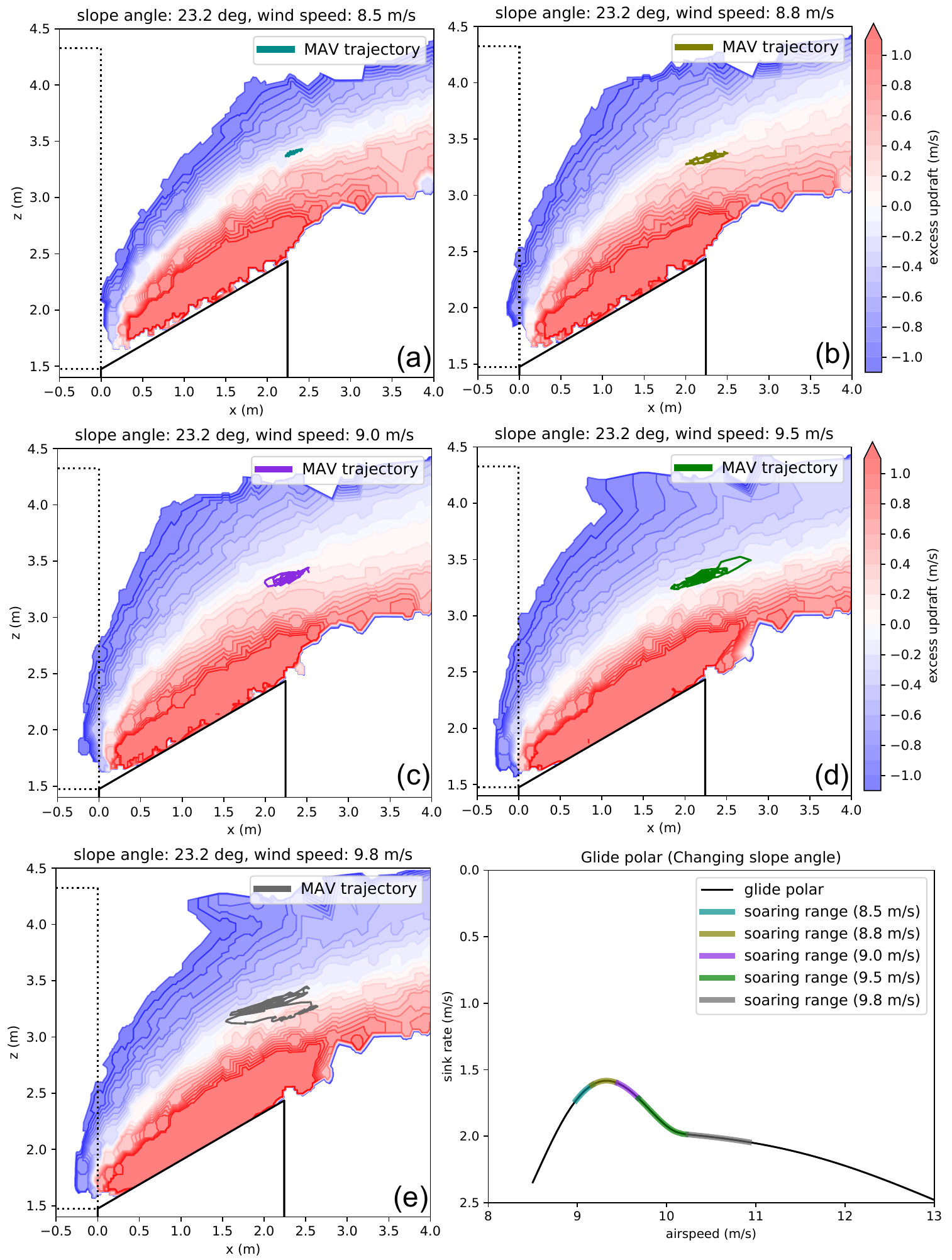}
 \caption{Case 1: autonomous soaring in changing wind speed from 8.5 to 9.8 m/s. Contours of excess updraft and the MAV trajectory are shown for each nominal wind speed. In the last plot, the range of sink rate for each case is shown on the glide polar.}
 \label{f:ojf_test_contour_ws}
\end{figure}

\begin{figure}[tbhp]
\centering
 \includegraphics[width=1.0\linewidth]{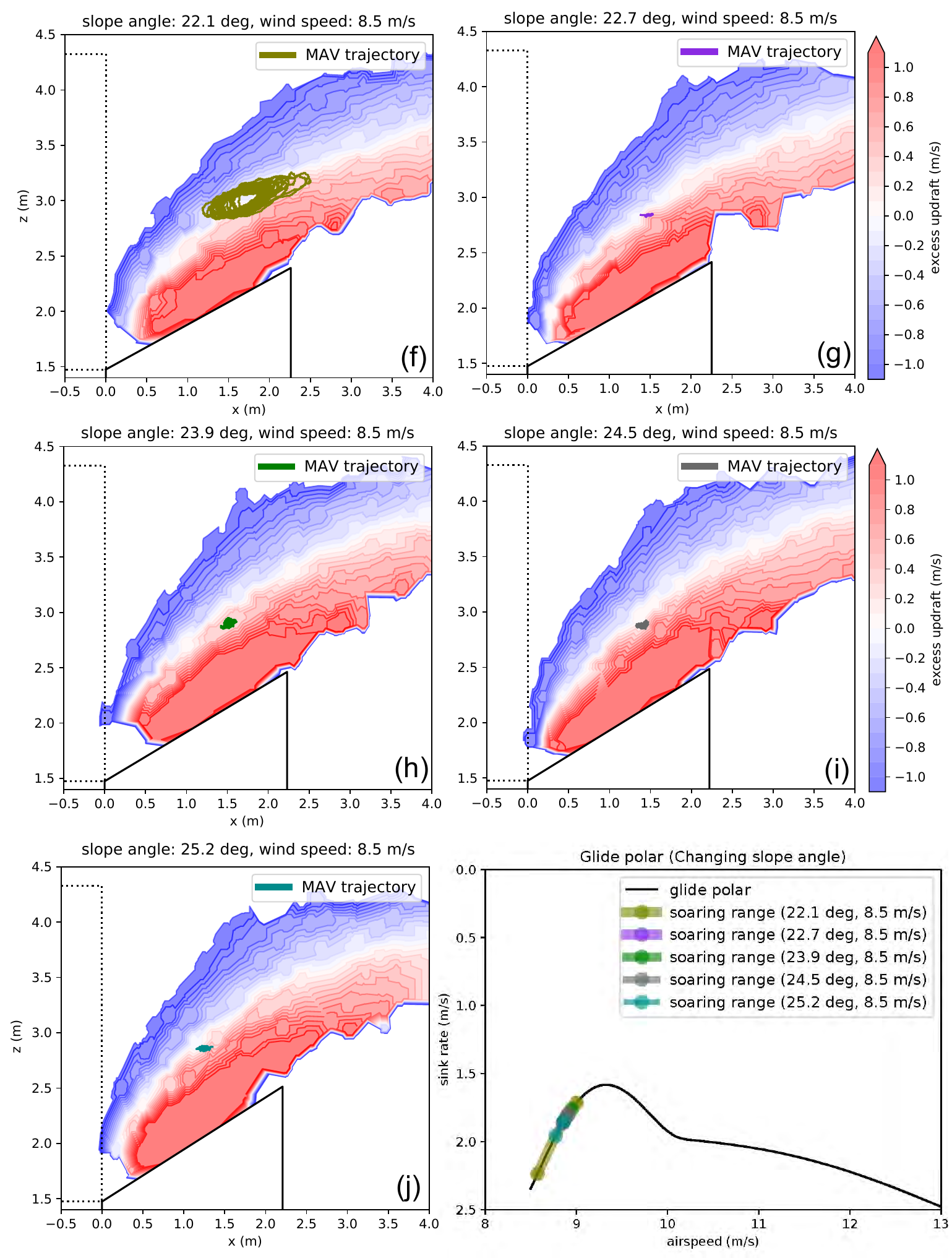}
 \caption{Case 2: autonomous soaring at varying slope angles from 22.1 to 25.2 degrees. The contours of excess updraft and the MAV trajectory for each slope angle are shown. On the last plot, glide polar with the range of the sink rate for each case is presented. The sink rates stayed in similar ranges because the nominal wind speed was fixed at 8.5 m/s in this case.}
 \label{f:ojf_test_contour_slope}
\end{figure}

For further analysis, we ran a CFD simulation for each combination of wind speed and slope angle.
Based on the wind field calculated by the CFD simulation and the MAV's sink rate, a feasible soaring region can be determined.
The MAV can soar where the updraft and sink rate are balanced (i.e. excess updraft = updraft - sink rate = 0), which is a white region in figures \ref{f:ojf_test_contour_ws} and \ref{f:ojf_test_contour_slope}.
The trajectory shows that the MAV mostly stayed in the region where the updraft and sink rate is balanced.
The plots show that the MAV generally resides inside or very close to the white areas, in which the predicted excess updraft is zero. The MAV does occasionally fly in slight non-zero excess updraft locations, which we expect to be due to imperfections in the predictions. These imperfections can have various sources: differences between the CFD simulation and the real world, imperfect sink rate measurement and fitting, airspeed sensor measurement, and the wind speed controller of the OJF also had steady-state errors within 0.1 m/s range.

%The trajectory of the MAV does not perfectly stay within the balanced updraft region. It is because there were various error sources: CFD simulation, sink rate measurement and fitting, airspeed sensor measurement, and the wind speed controller of the OJF also had steady-state error within 0.1 m/s range. Nonetheless, the flight log shows a good match with theoretical soaring region.

One thing we noticed is that the vertical position of the MAV went down as the wind speed increased.
Although this may seem counter-intuitive, the glide polar explains the behavior.
The minimum sink rate of the MAV occurs at around 9.1 m/s of airspeed. When the airspeed gets higher than that, the sink rate significantly increased. As a result, more updraft is required for the MAV to stay aloft, and the MAV lowers its altitude to find a stronger updraft.
%\textcolor{orange}{The MAV stayed near the end of the slope, not further back. It is considered that it is not favorable to fly further back due to higher turbulence intensity. This region requires more energy to soar because the MAV changes its attitude more aggressively to cope with the turbulence.}
In case of a varying slope angle, the position of the MAV moved forward as the inclination increased. It is because more updraft is generated with a larger slope angle, thus the position that balances the sink rate and the updraft moved forward. 

In both cases, the throttle usage and energy consumption decreased significantly after switching on the soaring mode.
We set a position where the MAV can hover using approximately 20\% of the throttle as a standby. During the soaring flight, the throttle usage dropped and remained at close to 0\% for almost all the time. The mean throttle usage of the entire soaring flight was 0.25\% for both cases, compared to 38\% for a nominal flight. There were a few moments that the MAV used throttle because it was necessary to avoid a stall or recover the position. 
Note that the size of the feasible soaring region is only 10 to 20 cm vertically, which is very small to maintain the position within the region even for a human pilot. Also, we changed the wind field during the flight so sometimes the MAV had to overcome a sudden change using the throttle.

% - 8.5 m/s soaring multiple trials -> different position but similar?nearby?
% - because of the starting point (standby) was almost the same every time
% - however it shows that it can soar in the (white) region.
% dynamic effect: theoretical soaring region
% higher than that: it sinks down because sink rate > updraft
% lower than the region: goes up bc sink rate < updraft
% but the x position does not have such thing...
% it almost always converged to similar position because we gave similar standby(starting) position.
% however it actually can soar at other positions too.
% at the back of the slope, turbulence intensity increases, so it requires more control effort than other positions, so the MAV does not want to go there.

% * dynamic effect?
% (sink rate <-> updraft & effect of the horizontal wind speed) 

\section{Conclusion} \label{conclusion}

%\textcolor{gray}{
%In this paper, we proposed an INDI controller with control allocation and a simulated-annealing-based soaring location searching method. The method was validated with autonomous orographic soaring flight tests in the wind tunnel.
%}

In this paper, we demonstrated the first autonomous orographic soaring in the real-world environment without a priori knowledge of the wind field, pre-defined trajectory planning, or manual initialization by a human pilot.
A fully autonomous orographic soaring was performed in the wind tunnel with changing environment settings.
With a combination of the local search algorithm and the INDI controller with control allocation, the MAV was able to soar autonomously with almost zero throttle in the updraft for over 25 minutes of flight time. Furthermore, we verified that the MAV can find a new soaring position when the updraft changes by using the proposed search method.

%\textcolor{orange}{The INDI controller is suitable for orographic soaring thanks to its robustness} and fast response, while control allocation enables the MAV to use throttle if necessary.
%\textcolor{orange}{
%The simulated-annealing-based searching method makes it possible to find a feasible soaring position even if the MAV is launched from an arbitrary position.
%The proposed method, AOSoar, makes it possible to perform autonomous soaring without a priori knowledge of the wind field.
%Furthermore, the AOSoar can cope with changing environmental conditions.}

For future work,
performing an outdoor flight test will be the next step. 
%Outdoor environments, especially around buildings, often have strong wind gusts. 
The proposed methods in this paper are applicable to outdoor flights because they do not depend on any pre-measured environmental condition. However, 
%more precise parameter tuning will be required due to the geometric and dimensional differences. 
%For the OJF test, the step size for searching for a suitable soaring location was set small because of the size of the cross-section of the wind tunnel. For outdoor tests, it should be increased or made adaptive for faster convergence. 
%Also, 
the MAV will have to be more aware of its environment, as the wind may change direction and it will have to sense and avoid obstacles during the search. Using additional sensors can be helpful for recognizing the surroundings.

\bibliographystyle{IEEEtran}
\bibliography{IEEEabrv,root}

\end{document}